\documentclass{article}
\usepackage{spconf,amsmath,epsfig}

\providecommand{\doi}[1]{doi: {\footnotesize \href{http://dx.doi.org/#1}{\path{#1}}}}

\usepackage[pdftex=true,breaklinks=true,hidelinks=true,colorlinks=true,citecolor=blue]{hyperref}

\usepackage[square,numbers]{natbib}

\title{Diffusion Models for Earth Observation Use-cases:\\from cloud removal to urban change detection}
\name{Fulvio Sanguigni$^{1,2}$, Mikolaj Czerkawski$^1$, Lorenzo Papa$^2$, Irene Amerini$^2$, 
B. Le Saux$^1$}
\address{\textsuperscript{1}$\Phi$-lab, ESRIN, European Space Agency,
         Frascati I-00044, Italy\\
         \textsuperscript{2}Department of Computer, Control and Management Engineering, AlcorLab\\
         Sapienza University of Rome, Via Ariosto 25 00185, Rome, Italy
         }

\begin{document}
%
\maketitle
\begin{abstract}
    The advancements in the state of the art of generative Artificial Intelligence (AI) brought by diffusion models can be highly beneficial in novel contexts involving Earth observation data. After introducing this new family of generative models, this work proposes and analyses three use cases which demonstrate the potential of diffusion-based approaches for satellite image data. Namely, we tackle cloud removal and inpainting, dataset generation for change-detection tasks, and urban replanning.
\end{abstract}
\section{Introduction}
\label{sec:intro}

    After a decade of Artificial Intelligence (AI) pervading all aspects of Earth observation~\cite{audebert2017deep} thanks to discriminative models being able to generate products out of satellite data such as classification maps, here comes the age of generative models able to directly encode the nature of Earth observation (EO) data. After Generative Adversarial Networks~\cite{mateogarcia2021GANsingeosciences, audebert2018generative} and Energy-based models~\cite{castillo2021energy}, the current state-of-the-art for generative models is largely achieved via denoising diffusion. The diffusion-based solutions gave rise to advancements such as text-to-image generators~\cite{ramesh2021zero,Rombach2022}, super-resolution models~\cite{saharia2022image}, or other image inverse tasks~\cite{saharia2022palette}. The high generative capability of these techniques could potentially be transferred for Earth observation tasks, however, the exact applications in this context are currently sparse.
    
    To provide a perspective on how these models could bring benefits for Earth observation data, several use cases are introduced and demonstrated herein. The three use cases cover a wide range of domains where diffusion models, and more specifically, diffusion-based inpainting, are applicable. This includes a use case of cloud removal already explored in the literature, a delivery of a synthetic change detection dataset, and finally, a use case of urban replanning with satellite imagery. In the following we present and explain Diffusion Models in Section~\ref{sec:background} then explore the three use cases in Section~\ref{sec:results} before drawing a few perspectives.

\section{Background: Diffusion Models}
\label{sec:background}
    In this work, the main focus is the image inpainting capability of denoising diffusion models. To outline the background of the employed techniques, a summary of the denoising diffusion generative approach as well as the variants thereof designed for inpainting is provided below.

    \subsection{Denoising Diffusion Generative Models} 
        Denoising diffusion generative models~\cite{sohl-dickstein15diffusion,Ho2020_nips} can overcome some common problems of earlier generative frameworks, such as the mode collapse and unstable trainign in GANs~\cite{goodfellow2014generative, mateogarcia2021GANsingeosciences} or suboptimal quality of synthesis in VAEs~\cite{kingma2014auto}.

        The generative process of denoising diffusion is based on a gradual transformation of a normal Gaussian sampling distribution $\mathcal{N}(0,I)$ to the distribution of data, by approximating the \textit{reverse process} using a deep neural network. The \textit{reverse process} is modelled as the reverse of the \textit{forward process} that transforms from the data distribution to a Gaussian distribution. The forward chain is obtained by iteratively degrading an image $\mathbf{x}_0$ from the data distribution for $T$ timesteps until reaching $\mathbf{x}_T$. The degradation is performed as additive Gaussian noise using a noise schedule $\beta_0,\beta_1...\beta_T$, with $\beta$ representing the variance of the noise injected into the image:
        
        \begin{equation}\label{eq:fw_qbeta}
            q(\mathbf{x}_t|\mathbf{x}_{t-1}) = \mathcal{N}(\mathbf{x}_t,\sqrt{1-\beta_t}\mathbf{x}_{t-1},\beta_t\mathbf{I})
        \end{equation}
        
        If $\alpha_t = 1-\beta_t$ and $\bar \alpha_t = \prod^T_{t=0} \alpha_t$, the corrupted image $\mathbf{x}_t$ at time step $t$ can be derived as $\mathbf{x}_{0}$:
        
        \begin{equation}\label{eq:fw:diffusion}
            \mathbf{x}_t = \sqrt{\bar\alpha_t}\mathbf{x}_0 + \sqrt{1-\bar\alpha_t}\mathbf{\epsilon}\quad\mathbf{\epsilon} \sim \mathcal{N}(0,\mathbf{I})
        \end{equation}

        The reverse operation cannot be expressed in closed form, and its distribution is instead approximated as $p_{\theta}(\mathbf{x}_{t-1}|\mathbf{x}_{t},t) = \mathcal{N}(\mathbf{x}_{t-1},\mathbf{\mu}_\theta,\Tilde{\beta_t}\mathbf{I})$, where:
        \begin{equation}
            \mathbf{\mu}_\theta = \frac{1}{\sqrt{\alpha_t}}\left(\mathbf{x}_{t-1} - \frac{1-\alpha_t}{\sqrt{1-\bar \alpha_t}}\epsilon_\theta \right)
        \end{equation}
        
        The estimation of the noise sample $\mathbf{\epsilon}_{\theta}$ can be obtained using a deep neural network with parameters $\theta$ trained using $x_t$ input and $\epsilon$ output derived from Equation~\ref{eq:fw:diffusion}. As detailed in~\cite{Ho2020_nips}, a reverse step can be computed as: 
        
        \begin{equation}\label{eq:bw}
            \mathbf{x}_{t-1} = \frac{1}{\sqrt{\alpha_t}}\left(\mathbf{x}_t-\frac{1-\alpha_t}{\sqrt{1-\bar\alpha_t}}\mathbf{\epsilon}_{\theta}(\mathbf{x}_t,t)\right) + \sigma_t \mathbf{z}
        \end{equation}

        A new sample can be generated by deriving consecutive samples of the diffusion chain, starting from $t=T$ (a sample derived from pure noise), and iterating towards $t=0$, which arrives at the original data distribution. The exact number of steps $T$ is a hyperparameter, and in the standard definition of the diffusion probabilistic models~\cite{Ho2020_nips}, can be expected to be in the range from several hundred to several thousand. This makes diffusion models generally more computationally expensive than earlier models, like GANs~\cite{goodfellow2014generative} or VAEs~\cite{kingma2014auto}.
        
        In the domain of Earth Observation, the application of diffusion models has so far been moderate and involved tasks such as cloud removal ~\cite{zhao2023cloud} or controllable image synthesis ~\cite{controllable_rs}. The use cases proposed in this work aim to motivate further work on this topic and demonstrate the diversity of potential impact.
    
    \subsection{Inpainting with Denoising Diffusion Models}

        Inpainting can be performed in at least two distinct ways based on diffusion models. One variant, RePaint~\cite{Lugmayr2022} is based on a masked mixing operation applied to the denoised signal, where an unconditional generative model can be employed for the task. The second variant requires expansion of the denoising network input channels to accommodate the extra condition of the inpainted image and the corresponding mask.

        Individual use cases presented in this work demonstrate both variants.

        \begin{figure}[h]
            \centering\includegraphics[width=0.5\textwidth]{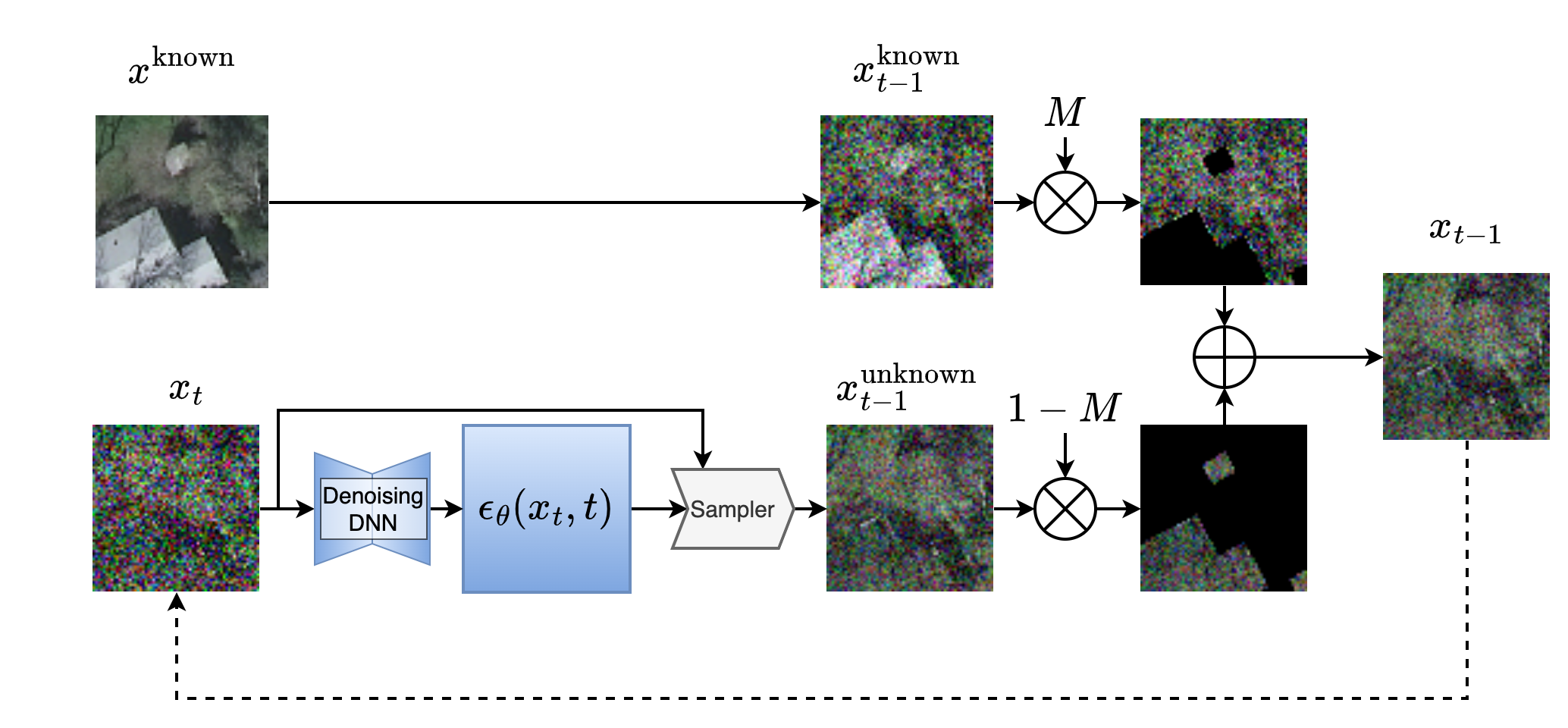}
            \caption{Inpainting Variant 1: RePaint \cite{Lugmayr2022}, where a partially known image $x^{\textrm{known}}$ is mixed with the generated sample $x_t$ using a mask $M$ to produce an inpainting at a time step $t-1$.}
            \label{fig:repaint}
        \end{figure}
        
        \subsubsection{Variant 1: RePaint}

            The RePaint technique~\cite{Lugmayr2022} allows for the reuse of an unconditional denoising diffusion model by mixing the denoised generated image with the input masked image $x^{\textrm{known}}$. As shown in Figure~\ref{fig:repaint}, this can be implemented via mixing the  generated image $x_{t-1}$ with the noisy input masked image $x_{t-1}^{\textrm{known}}$ (passed through the forward chain until $t-1$) using the mask $M$ at each step of the reverse chain.

            In this work, a version of RePaint with modifications is used, where we do not deploy further steps inside every timestep iteration.

        \subsubsection{Variant 2: Concatenation}

\begin{figure}
                \centering
                \includegraphics[width=\columnwidth]{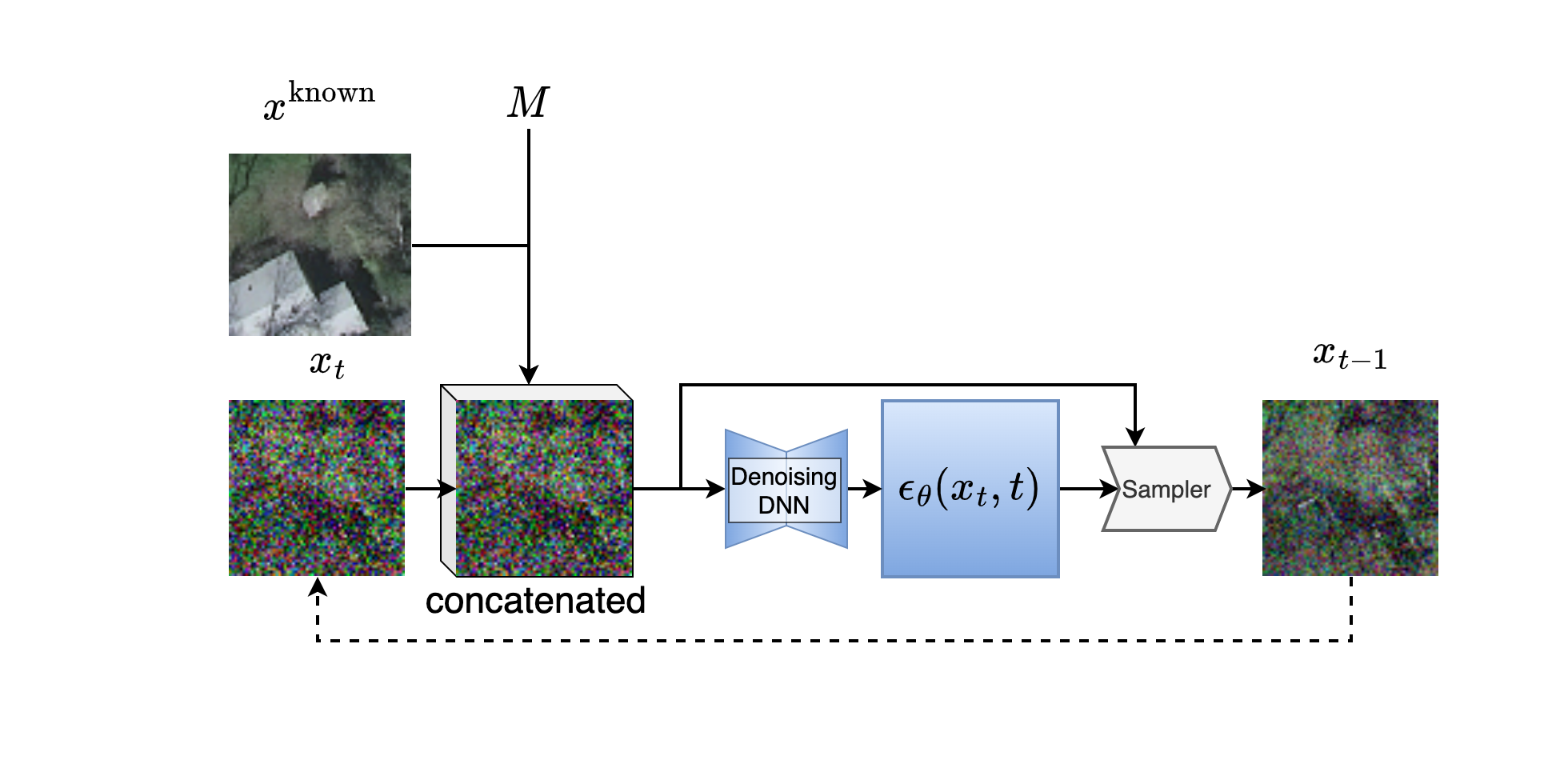}
                \caption{Diagram of the concatenation inpainting approach for diffusion models.}
                \label{fig:concat-diagram}
            \end{figure}

            Another popular approach for diffusion-based inpainting relies on input concatenation, as illustrated in Figure~\ref{fig:concat-diagram}, where the denoising network ingests the diffused signal $x_t$, but also the known image $x^{\textrm{known}}$ and the inpainting mask $M$. Unlike the RePaint variant described above, this requires a new network architecture to be defined and trained. This increased cost, however, gives an opportunity to create a model optimized directly for the inpainting task. This approach has been commonly used in LatentDiffusion~\cite{Rombach2022} and StableDiffusion~\cite{Rombach2022} models (albeit in latent space, rather than direct image space), and in this work, an inpainting StableDiffusion model is used to generate text-conditioned samples in the urban replanning use case.

\section{Exploring EO Use-Cases}
\label{sec:results}
    Three use cases, all taking advantage of the inpainting capability of diffusion models are described below to demonstrate the diversity of potential areas of impact in the field of Earth observation.
    
    \subsection{Inpainting for Cloud Removal}
        \begin{figure}[h]
            \centering
            \setlength\tabcolsep{2 pt}
            \begin{tabular}{ccc} 
                Input & Mask & Output
                \\
                \includegraphics[width=0.31\columnwidth]{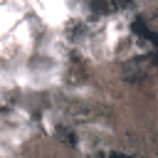} & \includegraphics[width=0.31\columnwidth]{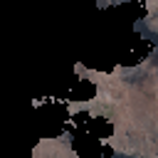} & \includegraphics[width=0.31\columnwidth]{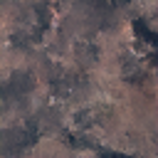} \\
                \includegraphics[width=0.31\columnwidth]{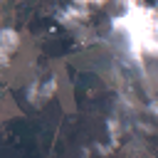} & \includegraphics[width=0.31\columnwidth]{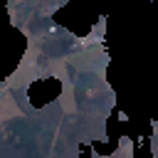} & \includegraphics[width=0.31\columnwidth]{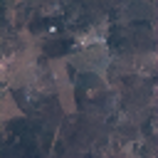} \\
            \end{tabular}
            \caption{Selected samples computed on the cloud removal task. Note that these samples of real clouds come from a different set than our reported numerical results}
            \label{fig:cloud_inpaint}
        \end{figure}
        
        The presence of clouds in satellite images poses a challenge both for visual inspection and for downstream tasks such as land cover classification. This is particularly problematic for critical real-time applications, where it is not feasible to wait for new acquisitions.
        
       Demonstrated below is how an unconditional diffusion model can be utilised for the task of cloud removal. A custom model has been trained on a set of about 8,000 Sentinel-2 images (RGB bands only in 64x64 resolution) for 250 epochs with 128 of batch size, and then used with the RePaint variant~\cite{Lugmayr2022} described earlier. This has been tested on the Sentinel-2 Cloud Mask Catalogue \cite{francis_alistair_2020_4172871}, a Sentinel-2 dataset containing cloud masks. Applied as an inpainting scheme by reusing random masks applied on the cloud-free images to ensure access to ground truth, the proposed model achieved an SSIM of 0.691 and PSNR of 24.593 dB. Inpainted examples from samples with real clouds are shown in Figure~\ref{fig:cloud_inpaint} where the dataset cloud masks were used for inpainting. The noise estimator is designed as a 4-layer U-Net with 128 base channels and a channel multiplier factor $f \in \{1,2,3,4\}$.

     \subsection{Generation of Change Detection Dataset}
        In EO there is a substantial need to label existing datasets to help current supervised approaches. Moreover, in some applications we have limited datasets in size, which lead to a poor supervised training.
        
        Another custom unconditional model has been trained on a remote sensing collection with the aim to generate a new dataset for change detection and consequently, provide more data for training on the downstream task. The model is used to generate a new pair of images, where the actual "change" is generated using our diffusion model  applying the inpainting approach we already used for cloud detection. The training is performed for 250 epochs with batch size 128. The neural network is a four-layer U-Net with 128 base channels and a channel multiplier factor $f \in \{1,2,3,4\}$. We train on the whole dataset cropped into $64 \times 64$ patches, with $50\%$ overlap between patches.
        
        To ensure the realism of the change regions (they are generally correlated with the overall structure of the image), the original masks and images from OSCD~\cite{daudt2018urban} are used and as a result, the original change is replaced with a synthetic one using the trained model. Due to the variational nature of the diffusion process, the resulting inpaintings are likely to introduce some change to the region.

        We make this dataset available in open access at the following \href{https://zenodo.org/record/8144238}{DOI:10.5281/zenodo.8144237} to motivate its use for supervised change detection pipelines.
        
        \begin{figure}[h]
            \centering
            \setlength\tabcolsep{2 pt}
            \begin{tabular}{ccc} 
                Input & Mask & Output
                \\
                \includegraphics[width=0.31\columnwidth]{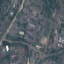} & \includegraphics[width=0.31\columnwidth]{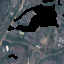} & \includegraphics[width=0.31\columnwidth]{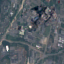} \\
                \includegraphics[width=0.31\columnwidth]{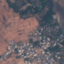} & \includegraphics[width=0.31\columnwidth]{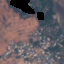} & \includegraphics[width=0.31\columnwidth]{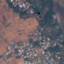}\\
            \end{tabular}
            \caption{Example images from the generated change detection dataset.}
            \label{fig:oscd_cd}
        \end{figure}
        
    \subsection{Urban Replanning}

        Another promising use case for the generative models proposed here relates to the urban replanning. With the ability to generate a wide range of realistic images, these models can serve as a visualization tool for communicating ideas about changes to urban environments. The capability to use text as an additional condition provides an interface for the user to control the process and specify the desired content of the generated image.

        This use case is demonstrated here by employing a StableDiffusion inpainting variant~\cite{Rombach2022} trained on samples from the LAION-5B dataset (which has been demonstrated to contain a considerable number of Earth observation images~\cite{czerkawski2023laion5b}) to recreate several scenes from the Inria Aerial Labeling dataset~\cite{maggiori2017can}, containing high-resolution aerial RGB images (30 cm) of detailed urban scenes. By masking out parts of the existing satellite images, the model can reconstruct alternative versions of the same image based on a provided text-prompt.

\begin{figure}[h]
            \centering
            \setlength\tabcolsep{2 pt}
            \begin{tabular}{ccc}
                Input & Mask & Output
                \\
                \includegraphics[width=0.31\columnwidth]{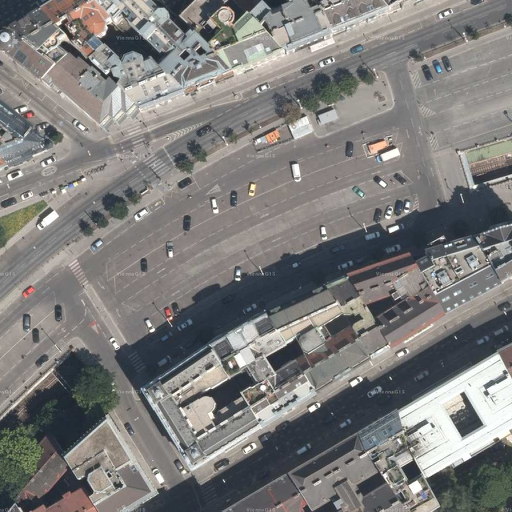} & \includegraphics[width=0.31\columnwidth]{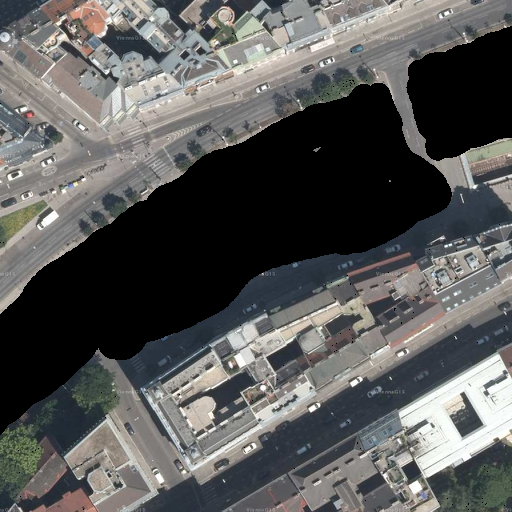} & \includegraphics[width=0.31\columnwidth]{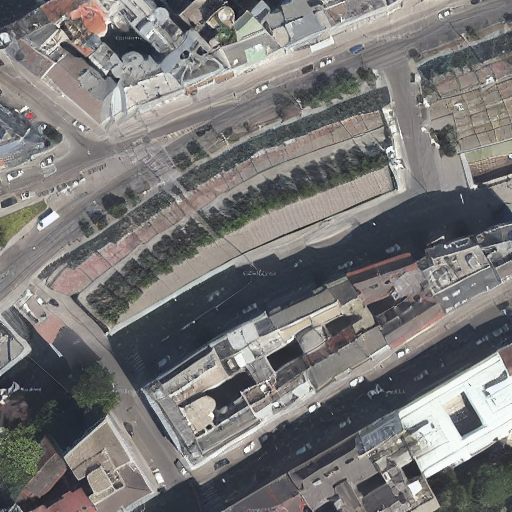} \\
                \includegraphics[width=0.31\columnwidth]{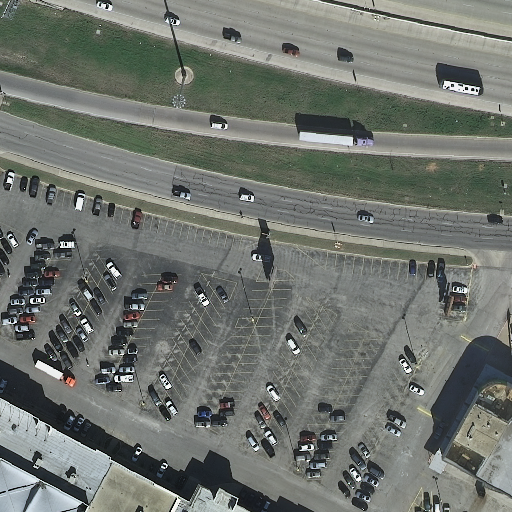} & \includegraphics[width=0.31\columnwidth]{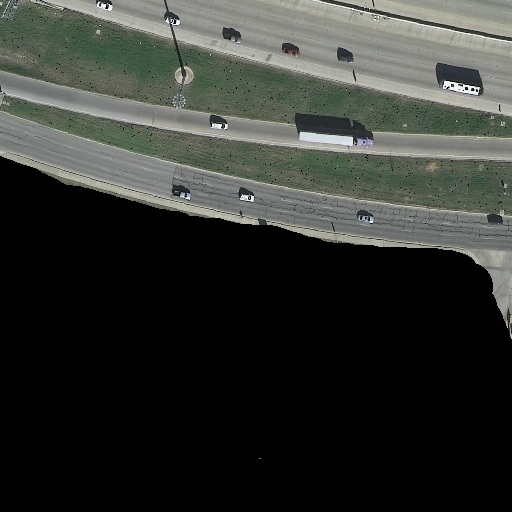} & \includegraphics[width=0.31\columnwidth]{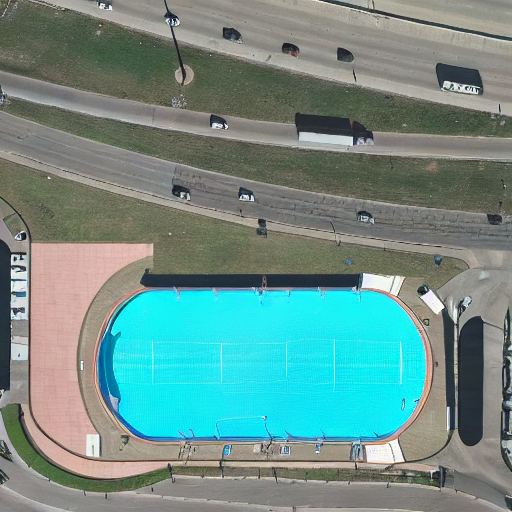} \\
            \end{tabular}
            \caption{Example of urban replanning visualization, where a car park is replaced with a pedestrianised area with trees (top row) and another car park is replaced with a large swimming pool (bottom row).}
            \label{fig:urban_replanning}
        \end{figure}

        Two examples of this use case are shown in Figure~\ref{fig:urban_replanning}, where a car park in Vienna (top row) is masked and replaced with a pedestrianised area and a car park in Austin (bottom row) with a large swimming pool. Transitions between real and generated parts are smooth and realistic thanks to diffusion.

        The application of diffusion-based models for urban replanning can lead to reduced costs of visualisation, where the model can support urban designers by generating images representing the potential ideas of rearranging urban spaces. The examples shown here were produced using an off-the-shelf model for image inpainting, which means that these results could potentially be improved further by training specialised models for this task.

\section{Conclusion}
\label{sec:conclusion}

    Diffusion models are a promising solution for generative modelling and hence, can support a wide range of use cases in the domain of Earth observation. The three presented examples demonstrate the value of diffusion models for a diverse set of tasks that involve satellite imagery. The inpainting capability of diffusion models has been verified by generating high-quality predictions for the tasks of cloud removal, dataset generation, and urban replanning, with the hope that it inspires the introduction of further remote sensing applications of diffusion models for the benefit of other downstream tasks.


\bibliographystyle{plainnat}
\bibliography{main.bbl}

\small

\end{document}